\documentclass[a4paper]{scrartcl}

\AtBeginDocument{%
  \providecommand\BibTeX{{%
    \normalfont B\kern-0.5em{\scshape i\kern-0.25em b}\kern-0.8em\TeX}}}

\usepackage{authblk}
\usepackage{algorithm}
\usepackage{algpseudocode}
\usepackage{xspace}
\usepackage{dcolumn}
\usepackage{subcaption}

\usepackage{graphicx}
\usepackage{amsmath,amsfonts,amssymb,amsthm}
\usepackage{natbib}
\usepackage{hyperref}
\usepackage{booktabs}

\graphicspath{{figures/}}

\def \mlflow {0b0df16d83f2466698edc31420951a85}

\newcommand{\Reals}{\mathbb{R}}

\newcommand{\matr}[1]{\ensuremath{\boldsymbol{\mathrm{#1}}}}
\newcommand{\diag}[1]{\ensuremath{\boldsymbol{\mathrm{diag}}\left(#1\right)\xspace}}

\newcommand{\Embedding}{\matr{E}^{(n)}}
\newcommand{\Key}{\matr{K}}
\newcommand{\Query}{\matr{Q}}
\newcommand{\Value}{\matr{V}}
\newcommand{\Mask}{\matr{M}}
\newcommand{\Sequence}{\matr{S}}
\newcommand{\Grid}{\matr{G}}
\newcommand{\Attention}{\matr{A}}
\newcommand{\Weight}{\matr{W}}
\newcommand{\Result}{\matr{R}}
\newcommand{\Normalizer}{\matr{N}}
\newcommand{\Perm}{\matr{P}}

\newcommand{\row}{\ensuremath{\mathrm{row}}}
\newcommand{\col}{\ensuremath{\mathrm{col}}}

\newcommand{\gamecontext}{\ensuremath{\mathrm{game-context}}}
\newcommand{\teamstrength}{\ensuremath{\mathrm{team-strength}}}
\newcommand{\playerstrength}{\ensuremath{\mathrm{player-strength}}}

\newcommand{\game}{\ensuremath{\mathrm{game}}}
\newcommand{\team}{\ensuremath{\mathrm{team}}}
\newcommand{\player}{\ensuremath{\mathrm{player}}}

\renewcommand{\vec}[1]{\boldsymbol{\mathrm{#1}}}

\newcommand{\embedding}{\vec{e}}
\newcommand{\ones}{\vec{1}}

\DeclareMathOperator{\attention}{Attention}
\DeclareMathOperator{\softmax}{Softmax}
\DeclareMathOperator{\axialattention}{AxialAttention}

\newtheorem{observation}{Observation}[section]

\begin{document}

\title{Large-Scale In-Game Outcome Forecasting for Match, Team and Players in Football using an Axial Transformer Neural Network}

\author{Michael Horton}

\author{Patrick Lucey}
\affil{Stats Perform Inc., Chicago, IL, USA 60611}


\maketitle

\begin{abstract}
  Football (soccer) is a sport that is characterised by complex game play, where players perform a variety of actions, such as passes, shots, tackles, fouls, in order to score goals, and ultimately win matches.
  Accurately forecasting the total number of each action that each player will complete during a match is desirable for a variety of applications, including tactical decision-making, sports betting, and for television broadcast commentary and analysis.
  Such predictions must consider the game state, the ability and skill of the players in both teams, the interactions between the players, and the temporal dynamics of the game as it develops.

  In this paper, we present a transformer-based neural network that jointly and recurrently predicts the expected totals for thirteen individual actions at multiple time-steps during the match, and where predictions are made for each individual player, each team and at the game-level.
  The neural network is based on an \emph{axial transformer} that efficiently captures the temporal dynamics as the game progresses, and the interactions between the players at each time-step.  
  We present a novel axial transformer design that we show is equivalent to a regular sequential transformer, and the design performs well experimentally.
  We show empirically that the model can make consistent and reliable predictions, and efficiently makes $\sim$75,000
  live predictions at low latency for each game.
 \end{abstract}

\begin{figure*}
  \centering
  \includegraphics[width=0.8\linewidth,page=12]{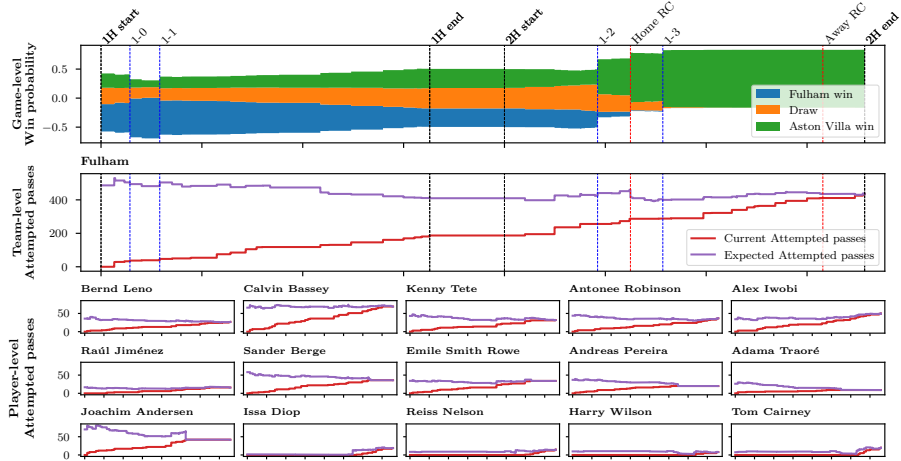}
  \caption{Real-time forecasts at game, team and player level for Fulham (1) vs. Aston Villa (3) on 19 October 2024.  As time progresses with the $x$-axis, the probability of each outcome is updated to account for the current game state.  A selection of key events indicated by dashed vertical lines, including goals (blue), red cards (red) and start and end of each half (black).}
  \label{fig:live-win-probability}
\end{figure*}



\section{Introduction}
\label{sec:introduction}
Football (soccer) is a game that can be viewed as a sequence of \emph{actions} such as passes, shots, tackles, fouls, etc., that are carried out by the players with the objective of scoring goals and ultimately winning the match.
The ability to accurately forecast---at a given point in the match---the total number of each such action that will be carried out by each player, and by each team, during the entire match is important for a variety of applications, 
such as: determining the starting lineup and substitutions; evaluating player and team performance; setting markets for sports betting; and supporting commentary and analysis during live broadcasts.


Until recently, match forecasting information was primarily obtained from betting markets~\cite[chapter~6]{graham-2024}, which can be viewed as machines for aggregating the ``wisdom of crowds'', where bettors vote their opinions through the liquidity they provide.
However, betting markets were only available on a limited number of outcomes such as game result and team goals scored, and were not available in-game.
Betting markets currently rely on human intervention, restricting their ability to scale to offering in-game player-level markets on multiple actions~\cite[chapter~5]{miller-2019}.
Subsequently, machine learning has facilitated data-driven forecasting models that address the scalability issue, however to-date these models have been limited in one or more ways: predicting a single target only, not making player-level predictions, or only making predictions pre-game.

Yet, there are many advantages to being able to make consistent in-game player-level forecasts on multiple target outcomes, such as the ability to analyse the impact of changes in game-state across all forecast outcomes, identify correlated forecasts, and to perform counter-factual analysis.
%
%
To that end, we present a single neural network model that jointly makes forecasts of the total number of actions made by each \emph{agent} (i.e. a player, a team or the overall game-state), for multiple actions, and at multiple time-steps during a match, see Figure~\ref{fig:live-win-probability}.

The model accepts
\emph{multi-modal} input features on all agents, and features that denote the a-priori strength and playing style of each player and team.
This model is characterized by a simple network structure, that still captures 
the interactions between the agents since the match started, and uses this shared state to forecast the final totals for each action by each agent, as well as the overall game outcome.

The model is based on the \emph{transformer}, a neural network component that has proven to be an effective and reliable learner across many domains, due to its apparent ability to capture interactions and dependencies across sequential, grid, and graphical data structures.
Moreover, it has been shown to effectively integrate and learn from the heterogeneous inputs inherent in multi-modal settings.
Initially designed for language modelling, the transformer is currently the backbone of state-of-the-art models in computer vision, vehicle motion forecasting, weather forecasting and drug discovery~\cite{grattafiori-2024,nayakanti-2023,price-2024,abramson-2024}. 

The actions that occur during a football match are the product of a complex system of interactions between players, teams and the game-state, and the task of predicting the total number of actions that will accrue to an individual agent over the duration of the match has several considerations that contextualise the prediction.
Consider the task of forecasting, at a particular point in time during the game, the total number of successful passes that a particular player will make during the match.
The prediction will be conditioned by several factors, including:

\begin{itemize}
\item The number of passes already made by the player.
\item The offensive strength of the player's team-mates, and the strategy the team is performing.
\item The defensive strength of the opposition team players, and their strategy.
\item The \emph{game state}, i.e.~the current score, remaining time, number of sent-off players, etc.
  \item The \emph{game context}, i.e.~the competition stage the game is player in e.g, whether it is a final in a knock-out tournament, or a meaningless end-of-season game.
  \item The expected time remaining until the player is substituted or the game ends.
  \item The \emph{momentum} of the current game state, e.g.~whether one team has been dominating the game-play recently.
\end{itemize}

\begin{figure*}
  \centering
  \includegraphics[width=1.0\linewidth,page=6]{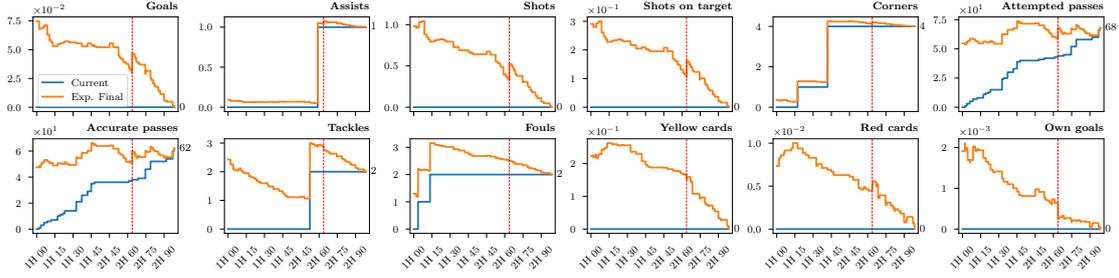}
  \caption{Live predictions for Youri Telemans (Aston Villa) for the 12 target actions during the Fulham vs. Aston Villa game, see Figure~\ref{fig:live-win-probability}.  For each target, the blue line represents the number of events that have already occurred, and the orange line is a live prediction of the total number of events that will occur during the game.  The predictions are sensitive to actions of other players, for example a red card was shown to an opposition player in the 62nd minute (indicated by the dashed vertical line).}
  \label{fig:player-live-predictions}
\end{figure*}

A model for this task must be \emph{consistent} in the predictions that it makes for the various actions.
The consistency may be trivial e.g., the number of shots on target by a particular player must be no greater than the total shots made,
or more complex, for example the total assists by all players in a team must be no greater than the total number of goals for the team.
The model should also be sensitive to the correlations between actions:
a team that is dominant will tend to make more passes and shots, and concede fewer fouls and yellow cards.
Ideally, the model has the capacity to make joint predictions on the game totals of all actions by learning the correlations and patterns between the actions.
Moreover, since the predictions are updated as the match progresses, the predictions must be temporally consistent in the sense that the predicted totals should never be less than the actual running totals, be smooth locally (with the exception of when significant events take place, such as goals), and converge to the actual totals at the end of the match.
The model should also be able to capture in-game dynamics, such as shifts in momentum, or the effects of goals, player dismissals, etc.\ as the rate that actions occur varies with the stage of the game, and with key events~\cite{dixon-1998}.
Figure~\ref{fig:player-live-predictions} is an example of action forecasts for a single player that illustrates the impact of an opponent being red-carded.

%

The required model thus has the following desiderata:
\begin{itemize}
\item Accepts inputs on a set of players with no implicit ordering, or requirement that the cardinality of the set is fixed.
\item Accepts inputs with different modalities, e.g. player, team, game-state, and both pre-game and in-game.
\item Makes consistent predictions for multiple actions for all agents.
\item Makes sequential predictions that are temporally consistent, and can capture in-game momentum and dynamics.
\end{itemize}

The presented neural network model, using only axial transformer layers and fully-connected linear layers, is able to fulfil these requirements.
The model is based on a novel formulation of the \emph{axial transformer} that efficiently integrates the temporal dynamics of the game as it progresses, and the interactions between players within each time-step in simple, efficient and principled approach.
The learned embeddings output by the final transformer layer are used to forecast the total of each action for each agent, by inputting them to a linear layer.

This model architecture is efficient: there are no duplicate features in the input; and the model makes multiple predictions at each time-step, currently it forecasts totals for each of twelve actions for each player in the match-day squad and for the two teams, and also predicts the game outcome at the game-state level, up to 505 predictions per time-step, and predictions are made temporally $\sim$150 time per match.
At inference time, the model achieves sub-second latency on commodity CPU hardware, and is thus appropriate for running in a real-time setting.

We are unaware of any existing work for the task of making such consistent real-time agent-level forecasts during matches from single model.
As such, the model was tested empirically and shown to obtain consistent performance across all the target actions.
Ablation tests were also performed that demonstrate that both player interactions and temporal dynamics contribute to fully capturing the current game state and subsequently making predictions.


The paper is arranged as follows.
We contextualise the work in Section~\ref{sec:related-work} by reviewing related work in sports prediction and the use of transformers for sequential and set-based prediction.
The data inputs and model are formally described in Section~\ref{sec:method} and empirical results are presented in Section~\ref{sec:experiments}.

\section{Related Work}
\label{sec:related-work}



\paragraph*{Match Forecasting}
\label{sec:match-forecasting}
Sports analysis is a research field that has developed significantly in recent years, enabled by a number of factors, including: the increased interest and commercialisation of professional sports and the related broadcasting and betting industries; the availability of detailed game data; and the relatively-constrained nature of games in terms of location and agents make it suitable for algorithmic analysis.
A particular area of research is \emph{forecasting}: predicting game outcomes, either prior to or during the game, see \citet{wunderlich-2020} for a review.

Historically, forecasting was restricted by the unavailability of detailed data to coarse targets, such as the outcome of a game~\cite{koopman-2019,wong-2025}, or the number of goals scored by each team~\cite{maher-1982,dixon-1997,lee-1997,karlis-2003}, and predictions were made \emph{pre-game}, typically based on the results of prior games.

The collection of fine-grained event tracking data in professional football allowed for predictions to be made on actions beyond just goals.  Event tracking is captured by human operators who log all action events, along with contextual attributes, such as the time, location, identity of involved players.\footnote{e.g.~\href{https://www.statsperform.com/opta/}{Opta~\nolinkurl{https://www.statsperform.com/opta/}},\\ \href{https://statsbomb.com/what-we-do/soccer-data/}{StatsBomb~\nolinkurl{https://statsbomb.com/what-we-do/soccer-data/}}}
These data allow for predictions to be made on actions other than just goals, and for each individual player.
Furthermore, event data allows for predictions to be made \emph{in-game}, where forecasts are made
for individual players on actions,
and are adjusted as the game proceeds and actions are recorded.
The event data may be further contextualised by combining it with player tracking data, so that the location and velocity of each player is known throughout the match~\cite{lucey-2015,chawla-2017,klemp-2021}.

Event and tracking has enabled models to make short-term predictions, such as estimating the type~\cite{simpson-2022}, or the value of the next event or events~\cite{lucey-2015,decroos-2019,fernandez-2021}.
Estimating the game outcome as the match proceeds, often known as \emph{live win probability}, is also possible with the game context extracted from event and tracking data~\cite{zou-2020,robberechts-2021,klemp-2021,rahimian-2024,yeung-2025}.

\paragraph*{Axial Transformer}
\label{sec:transformers}

The \emph{transformer} architecture~\cite{vaswani-2017} was initially designed for natural language processing (NLP) tasks such as machine translation, and was based on an encoder-decoder model.
Subsequently, the encoder component of the transformer was shown to be effective own its own on many NLP tasks, and that is was possible to leverage large unlabelled datasets in pre-training, and that the pre-trained model would transfer easily to directed tasks with minimal fine-tuning~\cite{devlin-2018}.
The transformer was subsequently shown to be effective in many other domains with structured sequential or grid-based data, including image recognition~\cite{dosovitskiy-2021}, drug discovery~\cite{jiang-2024,abramson-2024} and weather forecasting~\cite{price-2024,bojesomo-2024}
It is also used as the backbone of multi-modal architectures that fuse text and image inputs~\cite{kim-2021,wang-2022}.

The core operator in the transformer is the attention mechanism~\cite{bahdanau-2014} which operates over a single dimension, such as a sequence of word tokens in NLP tasks.
The \emph{axial transformer}~\cite{ho-2019} extended this to operate alternately over the rows and columns of pixels in an image.
The approach has subsequently been applied to learning the joint movement of agents over time~\cite{monti-2022,nayakanti-2023,xu-2023,jiang-2023}.


\paragraph*{Multi-task and Multi-modal Learning}
\label{sec:multi-task-learning}

Many approaches described above for action prediction problems are \emph{single-task} approaches, where predictions are made for a single action and a single player or team only, or for both teams at a time~\cite{karlis-2003}.
However, \emph{multi-task} learning, making simultaneous predictions on multiple targets, has been shown to improve the performance on all targets~\cite{caruana-1997}.
In the case of transformers, multi-task learning is routinely employed during pre-training~\cite{devlin-2018}, and has been shown to improve performance on downstream tasks~\cite{raffel-2019}.

Transformers have also been shown to be effective for \emph{multi-modal} learning, and 
have been used to integrate inputs of different modalities into a single model framework~\cite{mizrahi-2023,bachmann-2024}.

\section{Method}
\label{sec:method}

We present a model that accepts input features for different agents, both pre-game and over a time-series spanning a match, and uses a single transformer encoder backbone to learn embeddings that subsequently make predictions on multiple actions for each agent.
The model backbone is a series of axial transformer encoder layers, where each layer applies attention along both the temporal and agent dimensions.

\subsection{Preliminaries}
\label{sec:preliminaries}

The objective of the model is to predict the end-of-match totals for various actions for all involved agents.  Each \emph{target} is thus a discrete random variable $Y_{a, p, t}$, where $a$ is the action, $p$ is the agent and $t$ is the time-step .
Ideally, the model would predict the joint distribution of all targets $P(\mathbf{Y}) \colon \mathbf{Y} = \{Y_{a, p, t} | a \in \mathcal{A}, p \in \mathcal{P}, t \in \mathcal{T}\}$, where $\mathcal{A}$, $\mathcal{P}$ and $\mathcal{T}$ are the sets of actions, agents and time-steps, respectively.
However this is generally intractable as the space of possible outcomes is at least $2^{|\mathcal{A}| \times |\mathcal{P}| \times |\mathcal{T}|}$ (assuming that each $Y_{a,p,t}$ has only two outcomes).

Instead, we train a model to learn the marginal distribution for each target, $P(Y_{a, p, t}) = \sum_{\mathbf{Y} \setminus Y_{a, p, t}} P(\mathbf{Y})$, where $\mathbf{Y} \setminus Y_{a, p, t}$ is the set of all targets except $Y_{a, p, t}$.
Internally, the intention is that the model will approximate the joint distribution, and use this to determine the marginal distributions, and thus ensure consistency between the distributions for each target.

\paragraph*{Notation}
\label{sec:notation}

Matrices and tensors are denoted in boldface capitals, e.g. $\matr{M}$, and vectors as lowercase, e.g. $\vec{v}$.
Subscripts are used to distinguish objects, e.g. $\vec{v}_i$ or $\matr{M}_\row$.
Bracketed superscripts are used to distinguish entities with the same structure, e.g. input examples $\matr{X}^{(n)}$. 
Where ambiguity exists between e.g. matrix subscripts and element indices, indexing is denoted by the notation $(\matr{M})_{ij}$, which is the $i$-th element of the first dimension and $j$-th element of the second dimension, etc.
All enumerations used for indexing start with $1$.


\subsection{Data}
\label{sec:data}

The models is trained on a set of $N$ training examples $\mathcal{X}$.
The granularity of each training example is a single match, and for each match $n$ there are $T^{(n)}$ time-steps where an action event occurs.
There are also $P^{(n)}$ players in the match-day squad, and $2$ teams for each match $n$.

Each training example is a tuple $\left(\mathbf{X}^{(n)}, \mathbf{Y}^{(n)}\right)$, where $\mathbf{X}^{(n)}$ is the tuple of input tensors, and $\mathbf{Y}^{(n)}$ is the tuple of target tensors for match $n$.

\paragraph*{Input}
\label{sec:input}

The input tuple $\mathbf{X}^{(n)}$ contains the following tensors:
\begin{itemize}
\item $\mathbf{X}^{(n)}_\player \in \Reals^{P^{(n)} \times T^{(n)} \times D_\player}$ is the tensor of live player features, where $D_\player$ is the dimension of the player feature vectors.  The tensor contains indicator features for the player's position and team, and running totals of the twelve actions.
\item $\mathbf{X}^{(n)}_\playerstrength \in \Reals^{P^{(n)} \times D_\playerstrength}$ is the tensor of player strength features, with feature dimension $D_\playerstrength$.
  The player strength features are intended to capture the a-priori strength of the player, and are primarily aggregate statistics of the player's actions in recent games, such as the mean number of passes in the previous 5 games, the maximum fouls conceded in the previous 10 games, etc.
  In addition, there are features for the time since the previous match, distance from the player's home ground and distance from the previous match location, etc.
\item $\mathbf{X}^{(n)}_\team \in \Reals^{2 \times T^{(n)} \times D_\team}$ is the tensor of live team features, with feature dimension $D_\team$.
  The tensor contains indicator features for the team, and running totals of the actions already made by the team, similar to the live player features.
\item $\mathbf{X}^{(n)}_\teamstrength \in \Reals^{2 \times D_\teamstrength}$ is the tensor of team stren\-gth features, with feature dimension $D_\teamstrength$.
  The team strength features are primarily aggregate statistics of the team's actions in previous games, similar to the player strength features.
\item $\mathbf{X}^{(n)}_\game \in \Reals^{T^{(n)} \times D_\game}$ is the tensor of live game state features of dimension $D_\game$.
  The game state features contain attributes of the current event, such as the event-type, the game-clock time, and the event location.
\item $\mathbf{X}^{(n)}_\gamecontext \in \Reals^{D_\gamecontext}$ is the game context features vector of dimension $D_\gamecontext$.
  The event context features capture the context in which the game is played, such as indicator features for the competition (EPL, Champions' League, etc.), the hour the game is played at, etc.
\end{itemize}



An important attribute of this input scheme is that features are not duplicated.
the features are categorised by their \emph{modality}, such as: player-, team- or game-level; pre-game or time-step, and they are included in the corresponding tensor.
The model relies on the attention mechanism in the transformer layers to learn which features at each modality are relevant to predictions at other modalities.

\paragraph*{Target}
\label{sec:target}

The target tuple $\mathbf{Y}^{(n)}$ contains tensors for each of the modelled actions for each agent.
For a given action $a$, the target tensor contains the remaining number of actions for each agent in a tensor $\mathbf{Y}^{(n)}_{a} \in \Reals^{P^{(n)} \times T^{(n)}}$.
For example if $a$ is the number of shots, for player $p$ and for time-step $t$, $\left(\mathbf{Y}^{(n)}_{a}\right)_{pt}$ is the number of shots made by player $p$ in the interval $(t, T^{(n)}]$.
Using the remaining action count as ground-truth, rather than the total count, allows the count to be modelled using the Poisson distribution, which is a appropriate for many of the action targets considered here.

\subsection{Axial Transformer}
\label{sec:axial-transformer}

The axial transformer~\citep{ho-2019} is a key component of the model proposed in Section~\ref{sec:model}.  We propose a simple variation of the axial transformer that is simple, easy to implement using standard frameworks such as PyTorch, and we show that it is a computationally efficient implementation of the regular sequential masked self-attention~\cite{vaswani-2017}.

Following~\citet{vaswani-2017}, masked self-attention is an operation on a \emph{sequence} of embedding vectors contained in a matrix $\Sequence := [\embedding_1, \embedding_2,\dotsc,\embedding_s]^\intercal$ of length $S$:
\begin{align}
  \attention\left(\Key, \Query, \Value, \Mask\right) & := \softmax\left(\Attention\right) \Value   \label{eq:axial-attention} \text{, where} \\
  \Attention & = \dfrac{\Mask + \Key\Query^\intercal}{\sqrt{d}}
  \text{,}
\end{align}
where $\Key, \Query, \Value \in \Reals^{S \times D}$ are matrices computed from $\Sequence$ using a linear transformation (e.g. $\Key := \Sequence \matr{W_K}$ for some learnable $\matr{W}_k \in \Reals^{D \times D}$), and  $\Mask \in \Reals^{S \times S}$ is a \emph{mask matrix} that describes which elements in the sequence can be attended to.  The mask matrix is application-specific, and is constructed as $\Mask_{ij} = 0$ if $\Sequence_i$ can attend to $\Sequence_j$, and $-\infty$ otherwise.
The $\softmax$ operator is applied row-wise
using the element-wise operators exponential $\exp$ and Hadamard division:
\begin{align}
  \softmax(\Attention) & := \frac{\exp(\Attention)}{\vec{n}\ones^\intercal} \text{, where} \label{eq:softmax-defn} \\
  \vec{n} & = \exp(\Attention)\ones \text{, } \label{eq:normalizer}
\end{align}
where $\ones$ is the ones vector of dimension $S$, and the vector $\vec{n} \in \Reals^{s}$ is the vector containing the normalizing factor for each row
and thus normalizes each row of $\exp(\Attention)$ so that the each row sums to $1$.

Given an input $\Sequence$, the $\attention()$ operator returns $\Result \in \Reals^{S \times D}$, where each embedding vector $(\Result)_i$ is used to update the corresponding embedding vector $(\Sequence)_i$ in the subsequent layers of the transformer block.

\begin{figure}
  \centering
  \includegraphics[width=0.8\linewidth,page=4]{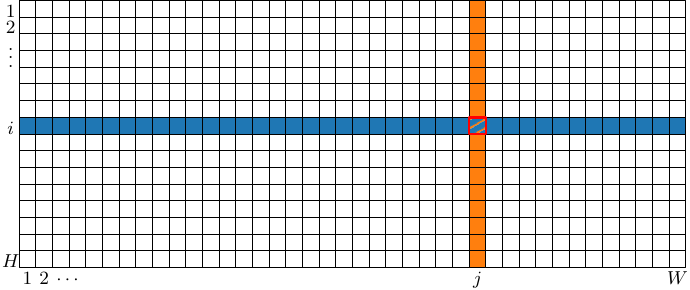}  
  \caption[Axial attention illustration]{Axial attention is used on a grid of embeddings.  For the given cell $(\Grid)_{ij}$ (with red border), self-attention is applied over the column $\Grid_{\cdot j}$ (yellow) and \textbf{masked} self-attention over the row $\Grid_{i \cdot}$ (green).}
  \label{fig:axial-attention-operation}
\end{figure}

In contrast to sequential attention, axial attention operates on a \emph{grid} of embedding vectors, contained in a $H \times W \times D$-dimensional tensor $\Grid$, where the $\attention()$ operator is applied to each row and to each column of the grid independently, obtaining the output tensors $\Result^\row, \Result^\col \in \Reals^{H \times W \times D}$, and then the outputs from these operations are aggregated.
The embedding vector $(\Grid)_{ij}$ is updated from the outputs of the $\attention()$ operation on row $i$ and column $j$, see Figure~\ref{fig:axial-attention-operation}.

Intuitively, it is desirable that the axial attention operation for updating $(\Grid)_{ij}$ should attend \textbf{equally} to both the embedding vectors in row $i$ and in column $j$.
This is achieved by simply adding $(\Result^\row)_{ij}$, i.e. the $j$-th element of the output of the row attention operation on row $i$, with $(\Result^\col)_{ij}$, the output of the $i$-th element of the output of the column attention on row $j$.
However, since the respective row and column $\softmax()$ operations have been individually normalized, we re-normalize over both using the row and column normalizing factors $(\vec{n}^\row)_i$ and $(\vec{n}^\col)_j$ respectively.
Naively, this approach will attend to the embedding vector $(\Grid)_{ij}$ twice, in both the row and column operation, however this can be avoided by asserting that this vector is masked in either the row or column $\attention()$ operation.
See Algorithm~\ref{alg:axial-attention} for the full $\axialattention$ operation.
\begin{algorithm}
  \caption{Axial Attention}\label{euclid}
  \label{alg:axial-attention}
  \small
  \begin{algorithmic}[1]
    \Require{$\Weight_Q, \Weight_K, \Weight_V$}\Comment{Weight matrices}
    \Procedure{AxialAttention}{$\Grid, \bigl\{\Mask^\row_i\bigr\}_{i=1}^{H}, \bigl\{\Mask^\col_j\bigr\}_{j=1}^{W}$}
    \Comment{Perform axial attention on grid, using mask and weights}
    \State $\Result^\row, \Result^\col \gets \emptyset, \emptyset$\Comment{3D tensors for results}
    \State $\Normalizer^\row, \Normalizer^\col \gets \emptyset, \emptyset$\Comment{Matrices for normalizing factors}
    \For{$i \gets 1, H$} \label{alg:axial-attention:row-start}
    \State{$\Sequence \gets (\Grid)_{i \cdot}$}\Comment{$\Sequence$ is $W \times D$ matrix}
    \State{$\Result, \vec{n} \gets \attention(\Sequence \Weight_Q, \Sequence \Weight_K, \Sequence \Weight_V, \Mask^\row_i)$},
    \State{$\Result^\row \gets \begin{bmatrix} \Result^\row \\ \Result \end{bmatrix}$}
    \State{$\Normalizer^\row \gets \begin{bmatrix}\Normalizer^\row \\ \vec{n} \end{bmatrix}$}
    \EndFor
    \For{$i \gets 1, W$} \label{alg:axial-attention:col-start}
    \State{$\Sequence \gets (\Grid)_{\cdot j}$}\Comment{$\Sequence$ is $H \times D$ matrix}
    \State{$\Result, \vec{n} \gets \attention(\Sequence \Weight_Q, \Sequence \Weight_K, \Sequence \Weight_V, \Mask^\col_j)$},
    \State{$\Result^\col \gets \begin{bmatrix} \Result^\col & \Result \end{bmatrix}$}
    \State{$\Normalizer^\col \gets \begin{bmatrix}\Normalizer^\col & \vec{n}\end{bmatrix}$}
    \EndFor
    \State{$\Result \gets \frac{\Normalizer^\row}{\Normalizer^\row + \Normalizer^\col} \odot \Result^\row + \frac{\Normalizer^\col}{\Normalizer^\row + \Normalizer^\col} \odot \Result^\col$} \label{alg:axial-attention:result} \Comment{Division is element-wise, $\odot$ is Hadamard product}
    \State{\Return $\Result$}\Comment{$\Result$ has same dimension as $\Grid$} 
    \EndProcedure
  \end{algorithmic}
\end{algorithm}

During a single $\axialattention()$ operation, $H$ row $\attention()$ operations on sequences of length $W$, and $W$ column $\attention()$ operations on sequences of length $H$, are performed.  Since the runtime and storage complexity of the $\attention()$ operation is quadratic in the sequence length, then the overall complexity is $O((H + W)HW)$.

In contrast to previous works using axial attention~\cite{ho-2019,wang-2022,nayakanti-2023} where the row and column attention operations are applied sequentially using distinct layers: $\Result = (\attention_\col \circ \attention_\row)(\Grid)$, here we add the weighted outputs of the row and column operations and share the attention weights between these operations.

\paragraph*{Equivalence to masked sequential attention}
\label{sec:seq-attn-equiv}
Next, we show that the additive axial attention operation can be reformulated to regular masked sequential attention with the following attention mask.

The grid of embedding vectors is ``unravelled'' to a row-major sequence of vectors $\Sequence = [ \Grid_{11}, \Grid_{12},\dotsc,\Grid_{1W},\Grid_{21},\dotsc,\Grid_{HW}]$.
Let $\Perm \in \{0,1\}^{HW \times HW}$ be the permutation matrix that reorders a column-major sequence to row-major, i.e. \[\Perm [\Grid_{11}, \Grid_{21},\dotsc, \Grid_{H1},\Grid_{12},\dotsc, \Grid_{HW}] = \Sequence\text{.}\]

The axial attention described above is equivalent to sequential attention on $\Sequence$ with the following attention mask:

\begin{align} \label{eq:seq-attn-mask}
  \Mask & := \Mask^\row + \Perm \left(\Mask^\col\right) \Perm^\intercal \text{, where} \\
  \Mask^\row & := \diag {\Mask^\row_1, \Mask^\row_2, \dotsc, \Mask^\row_H} \\
  \Mask^\col & := \diag{\Mask^\col_1, \Mask^\col_2, \dotsc, \Mask^\col_W}
\end{align}
and $\Mask^\row_i$ and $\Mask^\col_j$ are  the mask matrices for row $i$ and column $j$, respectively.
See Appendix~\ref{sec:axial-attention-proof} for a sketch proof of the equivalence.

This reformulation implies that any theoretical results, bounds, intuitions and heuristics that apply to regular masked self-attention will also apply to this formulation of axial attention.
Moreover, the time and storage complexity on this sequential self-attention operation is $O(H^2W^2)$, whereas axial attention is
sub-quadratic in the sequence length $HW$.

The axial attention operator is the core operator in the \emph{axial transformer} layer using a similar structure to the regular sequential transformer encoder layer. 
Axial attention is applied to the input grid, and this is followed by the standard feed-forward, layer-norm and skip connection components~\cite[Figure 1]{vaswani-2017}. 

The grid structure inherent in axial attention is appealing in the match forecasting setting; consider an embedding vector $(\Grid)_{ij}$, representing the state of a player $i$ at a particular time-step $j$.
In this case, the state contained in the embedding should be conditioned on all previous states for player $i$, and also on the states of all other agents in time-step $j$.
To update $(\Grid)_{ij}$, a row (temporal) attention operation $\attention(\Grid_i)$ is applied that attends only to previous embedding vectors $\{(\Result)_{ij'} \mid \forall j' < j\}$, which can be implemented using as the mask an upper-triangular mask matrix.
This allows the axial attention mechanism to capture the dynamics in the input features for each player over time, and thus can model changes in momentum, player fatigue, recency of significant in-game events, etc.
Similarly, a column (agent) attention operation $\attention(\Grid_{.j})$ attends to all embedding vectors in the column $\{(\Grid)_{i'j} \mid \forall i'\}$, thus permitting the axial attention mechanism to capture interactions between players.

\subsection{Model}
\label{sec:model}

\begin{figure}
  \centering
  \includegraphics[width=0.8\linewidth,page=2]{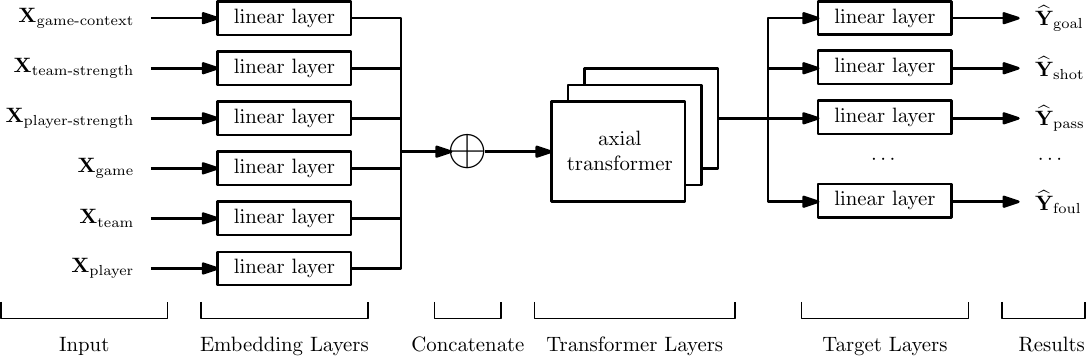}
  \caption{Overview of the neural network architecture.
    The tensors in the input are passed through the embedding layers to standardise the feature dimension, and then are stacked into a single tensor.
    This tensor is passed through the transformer layers, and the output embeddings are used to compute the outputs for each of the action targets.}
  \label{fig:model-architecture}
\end{figure}

The model is a multi-layer neural network consisting of three groups of layers, see Figure~\ref{fig:model-architecture}:
\begin{itemize}
\item \emph{Embedding layer} that maps the component tensors of the input into tensors with a common feature dimension, and concatenates them into a single input tensor.
\item \emph{Axial transformer layers} that performs attention along the temporal and agent dimensions.
\item \emph{Target layers} that map the output embeddings from the last transformer layer into tensors with the required feature dimension of each target metric.
\end{itemize}

\paragraph*{Embedding Layers}
\label{sec:embedding-layers}
The embedding layer is a linear transform on each input tensor that maps it to a tensor with a common feature dimension $D$.
For example, with the player input tensor the linear transform is a function
$f_{\player}: \Reals^{P^{(n)} \times T^{(n)} \times D_{\player}} \rightarrow \Reals^{P^{(n)} \times T^{(n)} \times D}$.
The output of the linear layer is a tuple of tensors, each with a common feature dimension, and so they may be concatenated along the temporal and agent dimensions to form a single tensor, in the following arrangement:
\begin{align*}
\Embedding_{0} & =
\begin{bmatrix}
f_{\gamecontext}\left(\matr{X}^{(n)}_{\gamecontext}\right) & f_{\game}\left(\matr{X}^{(n)}_{\game}\right) \\ f_{\teamstrength}\left(\matr{X}^{(n)}_{\teamstrength}\right) & f_{\team}\left(\matr{X}^{(n)}_{\team}\right) \\ f_{\playerstrength}\left(\matr{X}^{(n)}_{\playerstrength}\right) & f_{\player}\left(\matr{X}^{(n)}_{\player}\right)
\end{bmatrix} \\
& \in \Reals^{H \times W \times D}\text{, where } H = P^{(n)} + 2 + 1 \text{, and } W =  T^{(n)} + 1 \text{.}
\end{align*}

The embedding tensor $\Embedding$ is thus a grid of embedding vectors $\embedding_{ij} \in R^D$, where each embedding vector contains the information about the corresponding agent at a time-step.
Each row of $\Embedding$ contains a time-series for a particular \emph{agent} such, as a player, a team or the game-state.
Each column of $\Embedding$ is the set of embedding vectors for each agent at a particular time-step.

\paragraph*{Axial Transformer Layers}
\label{sec:transformer-layers}
The tensor $\Embedding_{0}$  is then passed through a series of $L$ axial transformer layers, as detailed in Section~\ref{sec:axial-transformer}.
%
Each layer $\ell$ accepts a tensor $\Embedding_{\ell-1}$
and outputs a tensor $\Embedding_\ell$ of the same dimension.
An autoregressive attention mask (i.e.\ a strict upper-triangular mask) is applied to the row (temporal) attention in each layer so that the embeddings in the current time-step $i$ can only attend to previous time-steps $[1..i-1]$.
The column (agent) attention allows agent $j$ in time-step $i$ to attend to all agents the current time-step.
The output $\Embedding_{L}$ of the final layer
has been updated by the transformer layers to integrate information from all input features up to and including the current time step.
This embedding is then used to directly make predictions for the required targets.

\paragraph*{Target Layers}
\label{sec:target-layers}

The final layer of the model is a set of linear layers that map the output embeddings $\Embedding_L$ of the final transformer layer to the required feature dimension of each target metric.
For each action target and modality (e.g. player, team or game), a linear layer is required that is shared over all agents and time-steps.
For example, if the target is the final goal count for a player, and we choose to model this as a Poisson distribution it is necessary to estimate a single parameter $\lambda$ for each player at each time step.
The linear layer is thus a function $h_a: \Reals^{P^{(n)} \times T^{(n)} \times D} \rightarrow \Reals^{P^{(n)} \times T^{(n)} \times 1}$.
Different distribution assumptions can be made and the only difference in the model is to the output parameter dimension of the corresponding linear layer.
We utilised several such assumptions, including Bernouilli, Poisson, log Gaussian, and ``model-free'' discrete distributions.

\subsection{Training}
\label{sec:training}

The corresponding loss function for the distribution assumption of each output target was chosen, e.g. Poisson negative log-likelihood for the Poisson distribution, binary cross entropy for the Bernouilli distribution, etc.
The losses were computed according to the ground truth value for each target in the training set, and the loss values were summed.
The model weights were updated from the total loss using the Adam optimizer~\cite{kingma-2014}, and the learning rate was adjusted on a schedule by cosine annealing, without warm restarts~\cite{loshchilov-2017}.

\subsection{Inference}
\label{sec:inference}

The model is designed to run in a live setting, where the invocation of the model is triggered by a key event occurring, of which there are 35 types, such as a foul, corner, goal, etc.
When a key event occurs, a message is triggered containing the attributes required to populate $\matr{X}^{(n)}_{\game}$, $\matr{X}^{(n)}_{\team}$, and $\matr{X}^{(n)}_{\player}$.
The contextual features for $\matr{X}^{(n)}_{\gamecontext}$, $\matr{X}^{(n)}_{\teamstrength}$ and $\matr{X}^{(n)}_{\playerstrength}$ are obtained from a relational database of historical game data, and are generated for the first (pre-game) event.
Once the features have been obtained they are stored in a key-value database, so for each event trigger, the contextual features and state features from all previous events are available.
The input is then assembled from the current contextual and previous features, and passed to the model.  The outputs are then enclosed in a data-structure that is sent to a messaging system for subsequent processing.


\section{Experiments}
\label{sec:experiments}

The model was evaluated using a dataset of 62,610 games from 28 competitions.  The dataset was split into a training set with games from 8 seasons from 2016--17 through to 2023--24 (58,501 games) and a test set with games from the partial 2024-25 season up until 15th December, 2024 (4,109 games).

The model makes approximately 505 predictions at each of the $\sim$150 time-steps per match, consisting of predictions on 12 target metrics for each of the $\sim$40 players (starting players and substitutes), and two teams, and a prediction for the game outcome.

The purpose of the experiments is twofold.
\begin{enumerate}
\item To evaluate the extent to which the model yields consistent results across all target actions, agents, and time-steps.
  Given the unavailability of existing baseline models or in-game player-level betting data, we examine how well \emph{calibrated} the model is.
\item To determine the importance of the model components to the quality of the results, we conduct several ablation tests to evaluate the importance of the inter-agent interactions, the temporal dynamics, and the pre-game context information.
  In addition, we compare the performance of the axial transformer presented here with the ``stacked'' axial transformer presented in~\citet{ho-2019}.
\end{enumerate}

In order to aggregate the performance over predictions for comparison, we use the following metrics:

\begin{description}
\item[Log-probability] The log-probability assigned to the ground-truth value from the estimated distribution provided by the model.  A simple average of all log-probabilities for a given agent (e.g. player or team) and target (e.g. goals, shots, passes) is computed.
\item[Calibration error] For a given agent and target, the modal value of the estimated distribution (i.e. the random variable that obtains the largest probability) is compared against the ground-truth value and the binary calibration is computed.
\end{description}



\subsection{Model Selection}
\label{sec:model-selection}

The models were trained on the training dataset of which 10\% of the examples were held back for validation.  Training occurred over $\sim$150,000 training-steps, and took approximately 15 hours running on a single NVIDIA A10 GPU.

Candidate models were created by varying the hyper-parameters used to train them, in particular the size of the latent dimension, the number of axial transformer layers, and the learning rate of the optimizer.
The data batch size was selected as the largest possible for the GPU memory limit.
For example, for ``our'' model, the configuration that obtained the best outcome, determined by the total validation loss, was configured with: a latent dimension of 128; 4 transformer layers; an optimizer learning rate of $0.0003$; and a batch size of $30$.

\begin{figure}
  \centering
  \includegraphics[width=0.8\linewidth,page=9]{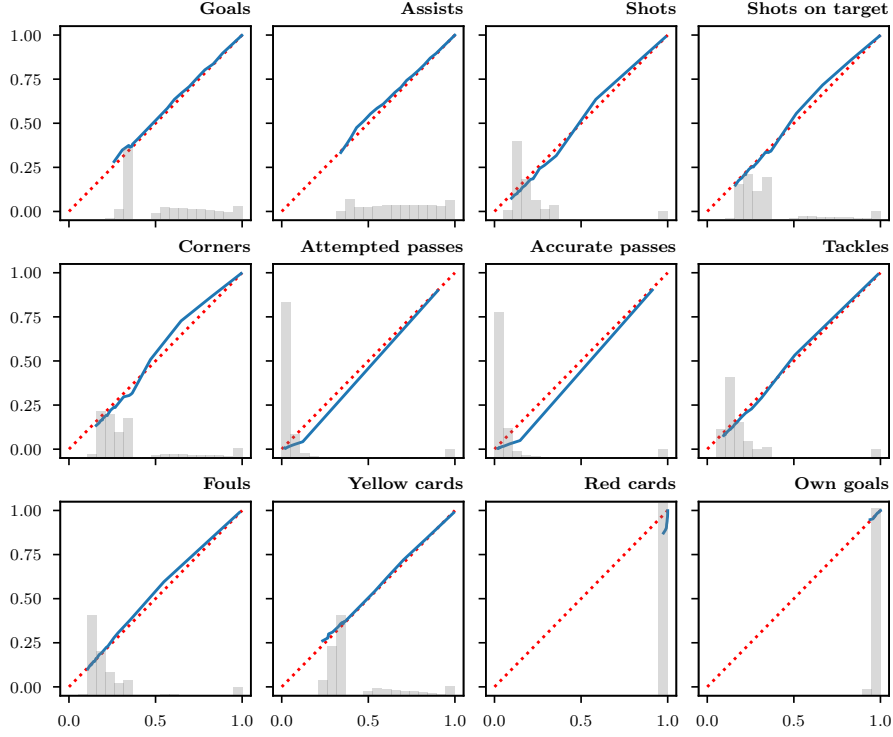}
  \caption{One-vs-one calibration plots for all targets, with the $x$-axis showing the predicted probability and the $y$-axis the actual probability.
    The blue line is the calibration, the red dashed line indicates the optimal calibration, and the grey histogram shows the density of predictions in each bin.}
  \label{fig:team-ovo-calibration}
\end{figure}

\subsection{Calibration}
\label{sec:calibration}


Figure~\ref{fig:team-ovo-calibration} contains the calibration curves for each of the action targets, using 20 bins.
In general, the model obtains well-calibrated predictions, however the calibration is weaker for targets where there are intervals that lack support, as shown by the grey histograms, e.g.~attempted and accurate passes, own-goals and red cards.

Red cards and own-goals are low probability events, and over the test set, the probability the model assigns to the zero category is no less than $0.84$ and $0.97$, respectively.
On the other hand, attempted and accurate passes are relatively frequent actions, and the range of possible values is large, and thus a relatively small probability is assigned to each value.
In both cases, the distribution of assigned probabilities as highly skewed.

\begin{table*}
\tiny
\caption{Calibration error and log-probability by each action target for each of the models in the ablation study.
  The best score across the models is marked in \textbf{bold}.
} 
\label{tbl:ablation-results}

\makeatletter
\newcolumntype{d}{D{.}{.}{-1} }
\newcolumntype{B}[3]{>{\boldmath\DC@{#1}{#2}{#3} }c<{\DC@end} } 
\makeatother

\newcommand{\best}[1]{\multicolumn{1}{B{.}{.}{-1}}{#1}}
\newcommand{\underflow}[1]{\ensuremath{{<}#1}}

\begin{tabular}{l *{5}{d}*{5}{d}}
\toprule
  & \multicolumn{5}{c}{Log-probability}
  & \multicolumn{5}{c}{Calibration error} \\
  \cmidrule(r){2-6} \cmidrule(l){7-11}
  & \multicolumn{1}{r}{\parbox[t]{0.8cm}{\raggedleft Ours}}
  & \multicolumn{1}{r}{\parbox[t]{0.8cm}{\raggedleft Ours w/o\\agent}}
  & \multicolumn{1}{r}{\parbox[t]{0.8cm}{\raggedleft Ours w/o\\temporal}}
  & \multicolumn{1}{r}{\parbox[t]{0.8cm}{\raggedleft Ours w/o\\pre-game}}
  & \multicolumn{1}{r}{\parbox[t]{0.8cm}{\raggedleft Ours w/\\stacked}}
  & \multicolumn{1}{r}{\parbox[t]{0.8cm}{\raggedleft Ours}}
  & \multicolumn{1}{r}{\parbox[t]{0.8cm}{\raggedleft Ours w/o\\agent}}
  & \multicolumn{1}{r}{\parbox[t]{0.8cm}{\raggedleft Ours w/o\\temporal}} 
  & \multicolumn{1}{r}{\parbox[t]{0.8cm}{\raggedleft Ours w/o\\pre-game}}
  & \multicolumn{1}{r}{\parbox[t]{0.8cm}{\raggedleft Ours w/\\stacked}}
  \\
  \midrule
  Goals             & \best{-0.111} & -0.114 & -0.124 & -0.124 & -0.112 & 0.002 & 0.013 & 0.004 & 0.014 & \best{0.002} \\
  Assists           & \best{-0.088} & -0.090 & -0.098 & -0.094 & -0.089 & 0.001 & 0.002 & 0.004 & 0.001 & \best{0.001} \\
  Shots             & \best{-0.515} & -0.532 & -0.575 & -0.597 & -0.520 & \best{0.021} & 0.022 & 0.028 & 0.044 & 0.021 \\
  Shots on target   & \best{-0.260} & -0.266 & -0.289 & -0.296 & -0.262 & 0.016 & 0.018 & \best{0.011} & 0.031 & 0.015 \\
  Corners           & \best{-0.289} & -0.299 & -0.323 & -0.320 & -0.292 & 0.019 & 0.020 & \best{0.013} & 0.030 & 0.017 \\
  Attempted passes  & \best{-3.039} & -3.333 & -3.501 & -3.735 & -3.062 & \best{0.074} & 0.093 & 0.093 & 0.102 & 0.076 \\
  Accurate passes   & \best{-2.765} & -3.002 & -3.199 & -3.403 & -2.781 & \best{0.075} & 0.094 & 0.096 & 0.109 & 0.078 \\
  Tackles           & \best{-0.608} & -0.620 & -0.674 & -0.661 & -0.614 & 0.027 & 0.025 & 0.029 & 0.028 & \best{0.024} \\
  Fouls             & \best{-0.517} & -0.530 & -0.569 & -0.558 & -0.524 & \best{0.018} & 0.023 & 0.021 & 0.028 & 0.020 \\
  Yellow cards      & \best{-0.202} & -0.209 & -0.220 & -0.211 & -0.203 & 0.019 & 0.020 & \best{0.004} & 0.023 & 0.019 \\
  Red cards         & \best{-0.018} & -0.019 & -0.020 & -0.019 & -0.019 & \underflow{0.001} & \best{\underflow{0.001}} & 0.003 & \underflow{0.001} & \underflow{0.001} \\
  Own goals         & \best{-0.006} & -0.007 & -0.007 & -0.007 & -0.006 & \best{\underflow{0.001}} & \underflow{0.001} & 0.001 & \underflow{0.001} & \underflow{0.001} \\

  \bottomrule
\end{tabular}
\end{table*}

\subsection{Ablation Study}
\label{sec:ablation}

A key property of the model design is that it accepts and integrates the available game data using a principled approach, thus allowing joint prediction on all the targets for all the agents at all time-steps.
To evaluate the effectiveness of the model in integrating the available data, we conducted experiments where the model was modified to disable attention to different agents and time-steps.
The following ablated models were used:
\begin{description}
\item[Agent Attention] The temporal attention layers in the axial encoder were removed, so that
  the model would attend to other agents, but not temporally.
  For each agent, the pre-game context features were concatenated to the time-step features, so the available information remains the same.
\item[Temporal Attention] The agent attention layers in the axial encoder were removed, so that predictions for each agent were made without being able to attend to the other agents.
\item[Pre-game Context] The game context and team and player strength features were omitted from the model so that each in-game prediction was made with only reference to events that had occurred during the game.
\item[Stacked Axial Attention] The stacked axial attention mechanism from \citet{ho-2019} is used.
\end{description}

The summary results are shown in Table~\ref{tbl:ablation-results}.
Our model obtained smaller mean log-probabilities than all the ablated models, thus demonstrating that effectiveness of this implementation of the axial transformer.
For calibration, our model obtained the best score against a plurality of the targets, and was close to best for most of the remaining actions.
The exception is yellow cards, where the temporal ablation is best, however it was found that this classifier was simply predicting the majority class in all cases.






\section{Conclusion}
\label{sec:conclusion}

We present a model for forecasting the total number of multiple actions performed in football matches for all players, teams and for the game overall, and that can be made multiple times as the match progresses, $\sim$75,000 predictions per game.
The model is a novel variation of axial attention that computes a weighted sum of the row and column attention operations.
We show that this variation is equivalent to an instance of regular sequential self-attention, yet is computationally more efficient, and experiments show the this variation obtains better performance in comparison to the typical ``stacked'' axial attention.
The model was empirically evaluated by ablating different components, and the results showed that capacity of the model architecture to attend along both the temporal and agent dimensions, and the ability of the model to integrate different modalities of data, both contributed to the obtained performance.

\bibliographystyle{ACM-Reference-Format}
\bibliography{refs}

\appendix

\section{Axial Attention}
\label{sec:axial-attention-proof}

In Section~\ref{sec:axial-transformer}, we claim that an instance of masked axial attention is equivalent to an instance of regular masked sequential self-attention, where the grid $\Grid$ of embedding vectors is unravelled to the sequence $\Sequence$ and the attention mask is defined in Equation~\eqref{eq:seq-attn-mask}.
The following is a sketch proof of this claim, where we use this observation:

\begin{observation}
  \label{obs:softmax-pe}
  The $\softmax()$ operator, when applied to the attention matrix $\Attention$ is \emph{permutation equivariant} with respect to both row- and column-permutations, i.e. for any permutation matrix $\Perm$, we have:
  \begin{align*}
    \softmax(\Perm \Attention) & = \Perm \softmax(\Attention) \text{, and} \\
    \softmax(\Attention \Perm) & = \softmax(\Attention) \Perm \text{.}
  \end{align*}

\end{observation}
\begin{proof}
  The $\softmax$ operator is applied in a row-wise manner to the $\Attention$ matrix and ensures that each row sums to $1$.

  Thus the rows of the input can be reordered without altering the results in any row, and the result has a corresponding reordering.

  Furthermore, reordering the columns of the input does not change the resulting values, as the vector dot product and scalar division operations in Equation~\eqref{eq:softmax-defn} are both permutation invariant.  
\end{proof}

Recall from Section~\ref{sec:axial-transformer}, that the input grid of embedding vectors $\Grid \in \Reals^{H \times W \times D}$ is unravelled into a sequence $\Sequence \in \Reals^{HW \times D}$ in row-major format, and that the permutation matrix $\Perm \in \{0,1\}^{HW \times HW}$ reorders a sequence from column-major to row-major format.

We claim that the masked sequential self-attention operation~\cite{vaswani-2017} from Equation~\eqref{eq:softmax-defn} on $\Sequence$ using the mask matrix from Equation~\eqref{eq:seq-attn-mask}, is equivalent to the masked axial attention operation, detailed in Algorithm~\ref{alg:axial-attention}.

\begin{figure}
  \centering
  \includegraphics[width=0.8\linewidth,page=5]{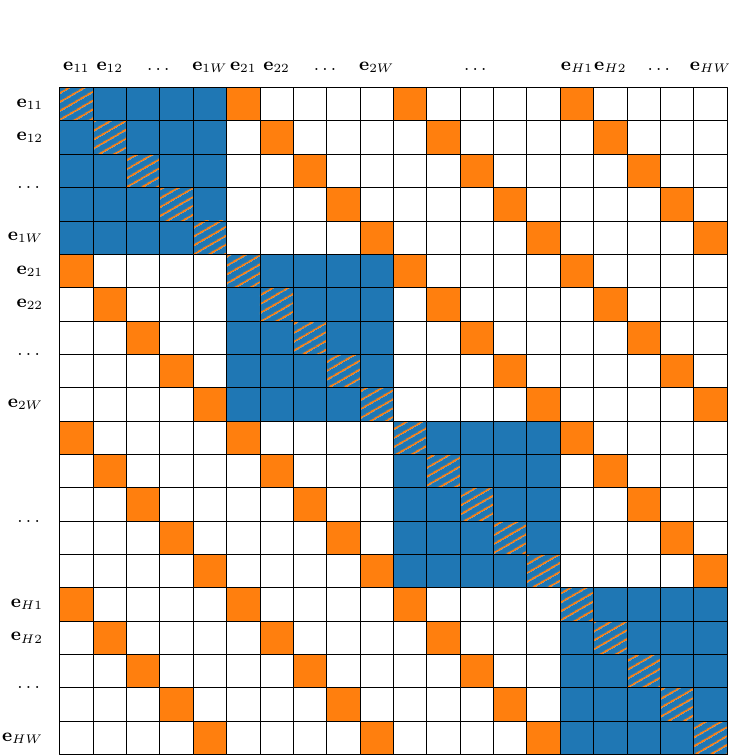}
  \caption{Illustration of mask matrix pattern on unravelled sequence $\Sequence$ for row (blue) and column (orange) axial attention, where colored cells are attended to (i.e. value of $0$) and blank cells are masked (value of $-\infty$).  If the diagonal elements $e_{ii}$ are masked in either of the mask matrices, then $\Mask^\row$ and $\Mask^\col$ are disjoint.}
  \label{fig:unravelled-mask}
\end{figure}

Consider the mask matrix from Equation~\eqref{eq:seq-attn-mask}.
The component $\Mask^\row$ ensures that (at most) an embedding vector will only attend to the vectors in the same row of $\Grid$ and, likewise, $\Mask^\col$ ensures that (at most) the embedding vector will only attend to the vectors in the same column of $\Grid$, see Figure~\ref{fig:unravelled-mask} for an illustration of the mask pattern.

Given the mask $\Mask$, it is sufficient to only compute the elements of the attention matrix $\Attention$ that are not masked.
For $\Mask^\row$, the unmasked components of $\Attention$ is the block diagonal matrix $\Attention^\row := \diag{\Attention^\row_1,\Attention^\row_2,\dotsc,\Attention^\row_H}$ where $\Attention^\row_i \in \Reals^{W \times W}$.
The unmasked components of $\Attention$ in $\Mask^\col$ are also a block diagonal matrix under row and column permutations, i.e. $\Perm (\Attention^\col) \Perm^\intercal$ where $\Attention^\col := \diag{\Attention^\col_1, \Attention^\col_2,\dotsc,\Attention^\col_W}$ where $\Attention^\col_j \in \Reals^{H \times H}$.

Using the pre-condition that either $\Mask^\row$ or $\Mask^\col$ masks the diagonal elements of $\Attention^\row$ or $\Attention^\col$ and thus $\Attention^\row$ and $\Attention^\col$ are disjoint, we have that $\Attention = \Attention^\row + \Perm (\Attention^\col) \Perm^\intercal$.
Furthermore, from Equation~\eqref{eq:axial-attention}, we can see that
\begin{align}
  \softmax(\Attention)\matr{V} = & \softmax(\Attention^\row + \Perm(\Attention^\col)\Perm^\intercal)\matr{V} \\
  \label{eq:attn-decomp:split-softmax} 
  = & \biggl(\frac{\vec{n}^\row}{\vec{n}^\row + \vec{n}^\col} \odot \softmax(\Attention^\row) + \nonumber \\
                                 & \frac{\vec{n}^\col}{\vec{n}^\row + \vec{n}^\col}\odot \Perm\softmax(\Attention^\col)\Perm^\intercal \biggr) \matr{V} \\
  \label{eq:attn-decomp:permute}
  = & \frac{\vec{n}^\row}{\vec{n}^\row + \vec{n}^\col} \odot \softmax(\Attention^\row) \matr{V} + \nonumber \\
                                 & \frac{\vec{n}^\col}{\vec{n}^\row + \vec{n}^\col} \odot \Perm\softmax(\Attention^\col) (\Perm^{-1} \matr{V}) \\
  = & \frac{\vec{n}^\row}{\vec{n}^\row + \vec{n}^\col} \odot \Result'_\row  + \frac{\vec{n}^\col}{\vec{n}^\row + \vec{n}^\col} \odot \Result'_\col \\
  = & \Result' \text{,} \label{eq:attn-decomp}
\end{align}
where $\Result' \in \Reals^{HW \times D}$ is the unravelled attention results in row-major layout.
Line~\ref{eq:attn-decomp:split-softmax} uses Observation~\ref{obs:softmax-pe}, and 
the terms are adjusted so that the attention weights from the $\softmax()$ operation sum to $1$.
In line~\ref{eq:attn-decomp:permute} we use the identity that $\Perm^{-1} \equiv \Perm^\intercal$ for any permutation matrix to reorder the rows of $\matr{V}$ to column-major.

Since $\Attention^\row$ and $\Attention^\col$ are block diagonal, multiplying by $\matr{V}$ and $\Perm^{-1}\matr{V}$ respectively is equivalent to multiplying each block by the corresponding sub-matrix of $\matr{V}$ and $\Perm^{-1}\matr{V}$, as is performed for-loops in lines~\ref{alg:axial-attention:row-start} and~\ref{alg:axial-attention:col-start} of Algorithm~\ref{alg:axial-attention}.
By ``re-ravelling``to the grid layout, we obtain an equivalent matrix to $\Result$ from line~\ref{alg:axial-attention:result} in Algorithm~\ref{alg:axial-attention}.



%

\section{Example Visualizations}

\begin{figure*}
  \begin{subfigure}{1.0\linewidth}
    \centering
    \includegraphics[width=\linewidth,page=5]{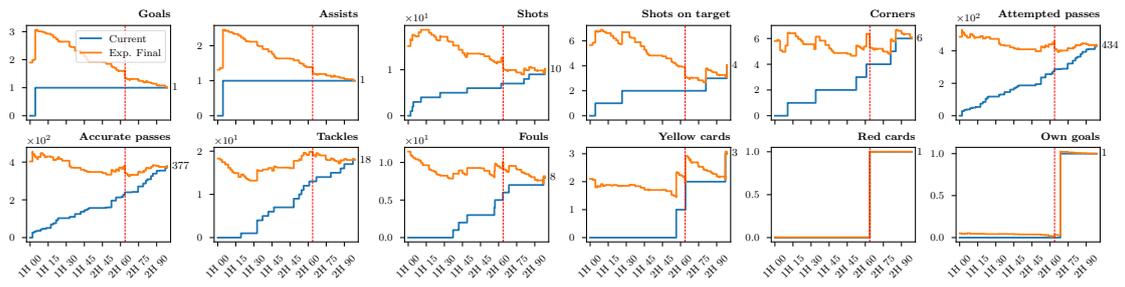}
    \subcaption{Fulham}
    \label{fig:time-series:team-1}
  \end{subfigure}
  \begin{subfigure}{1.0\linewidth}
    \centering
    \includegraphics[width=\linewidth,page=4]{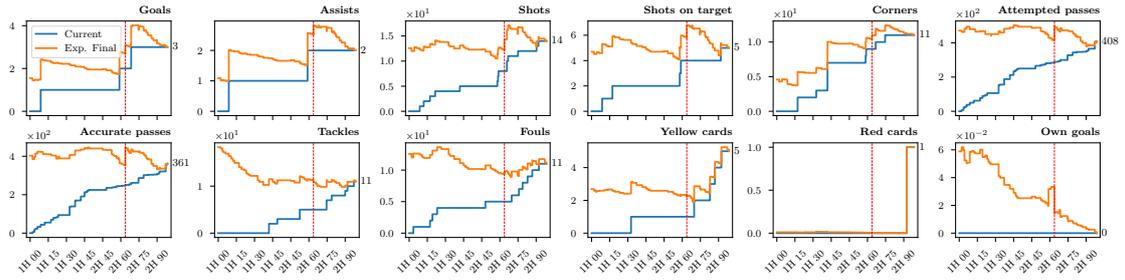}
    \caption{Aston Villa}
    \label{fig:time-series:team-2}
  \end{subfigure}
  \begin{subfigure}{1.0\linewidth}
    \centering
    \includegraphics[width=\linewidth,page=7]{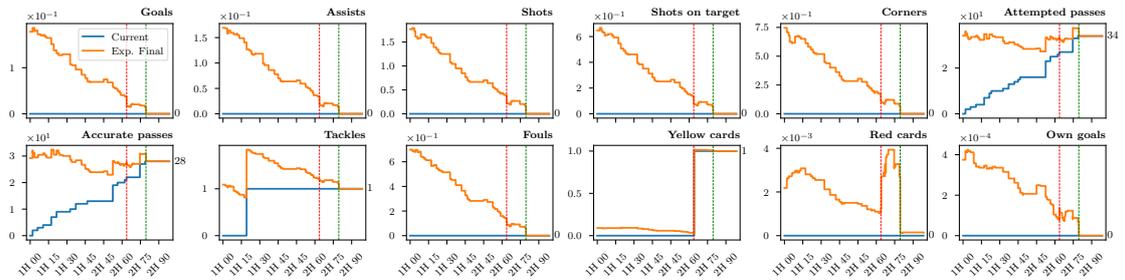}
    \caption{Emile Smith Rowe (Fulham)}
    \label{fig:time-series:player-1}
  \end{subfigure}
  \begin{subfigure}{1.0\linewidth}
    \centering
    \includegraphics[width=\linewidth,page=8]{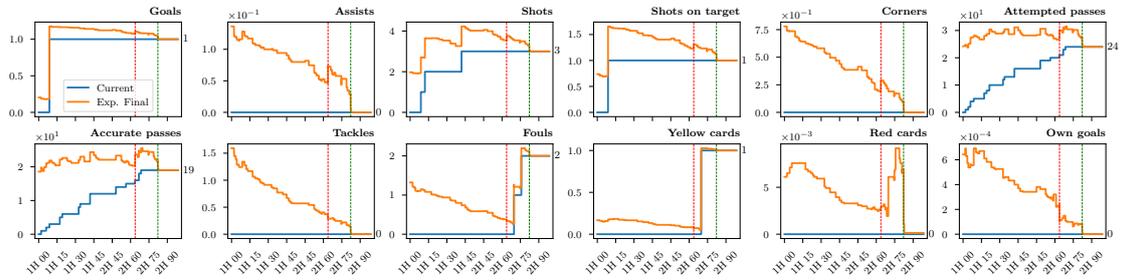}
    \caption{Morgan Rogers (Aston Villa)}
    \label{fig:time-series:player-2}
  \end{subfigure}
  \caption{Live forecasts for all target actions for Fulham, Aston Villa, and players Emile Smith Rowe (Fulham) and Morgan Rogers (Aston Villa), a match won 1--3 by Aston Villa.  The red dotted vertical line indicates the time-step that a Fulham player was given a red-card.
    Both Smith Row and Rogers are substituted, indicated by the green dotted vertical line.}
  \label{fig:time-series}
\end{figure*}

Figure~\ref{fig:time-series} are examples of predictions from the running example game Fulham vs. Aston Villa for both teams, and for one player from each team.
The red dotted vertical line indicates the time-step that a Fulham player is shown a red card and sent off.
At this point, the expected action totals for both teams and players adjust to this situation, e.g. fewer expected total shots for Fulham, and more for Aston Villa, and for own goals the expected action totals adjust in the opposite direction.

In Figures~\ref{fig:time-series:player-1}~and~\ref{fig:time-series:player-2}, the green dotted vertical line indicates the time-step when the player was substituted.
Note that for all actions, the predicted future actions is zero, with the exception of red and yellow cards, as substituted players may still receive cards, e.g. for dissent.

In Figure~\ref{fig:live-win-probability}, we can see that Fulham were favorites pre-game with a predicted win probability of 0.47.
At the start of the match, Fulham also have higher forecast totals for actions positively correlated with winning, such as goals, shots and assists, and lower forecasts for negatively correlated actions: fouls, yellow and red cards.
Furthermore, at the end of the match, Fulham as consistently under-performed the pre-game predictions, and has lost the match 1--3.  On the other hand, Aston Villa slightly over-performed their pre-game predictions, and won the match.

\end{document}